\documentclass[10pt,twocolumn,letterpaper]{article}

\usepackage{cvpr}
\usepackage{times}
\usepackage{epsfig}
\usepackage{graphicx}
\usepackage{amsmath}
\usepackage{amssymb}
\usepackage{algorithm}
\usepackage[noend]{algpseudocode}
\usepackage{enumitem}
\usepackage{multirow}      
\usepackage{multicol}

\usepackage[breaklinks=true,bookmarks=false]{hyperref}

\cvprfinalcopy 

\def\cvprPaperID{****} 

\ifcvprfinal\fi
\begin{document}

\title{\textit{AMC-Loss:} Angular Margin Contrastive Loss for Improved Explainability in Image Classification}

\author{Hongjun Choi, Anirudh Som, Pavan Turaga\\ \\
Geometric Media Lab\\
School of Arts, Media and Engineering, Arizona State University\\
School of Electrical, Computer and Energy Engineering, Arizona State University\\
{\tt\small hchoi71@asu.edu, asom2@asu.edu, pturaga@asu.edu}
}

\maketitle

\begin{abstract}
   Deep-learning architectures for classification problems involve the cross-entropy loss sometimes assisted with auxiliary loss functions like center loss, contrastive loss and triplet loss. These auxiliary loss functions facilitate better discrimination between the different classes of interest. However, recent studies hint at the fact that these loss functions do not take into account the intrinsic angular distribution exhibited by the low-level and high-level feature representations. This results in less compactness between samples from the same class and unclear boundary separations between data clusters of different classes. In this paper, we address this issue by proposing the use of geometric constraints, rooted in Riemannian geometry. Specifically, we propose Angular Margin Contrastive Loss (AMC-Loss), a new loss function to be used along with the traditional cross-entropy loss. The AMC-Loss employs the discriminative angular distance metric that is equivalent to geodesic distance on a hypersphere manifold such that it can serve a clear geometric interpretation. We demonstrate the effectiveness of AMC-Loss by providing quantitative and qualitative results. We find that although the proposed geometrically constrained loss-function improves quantitative results modestly, it has a qualitatively surprisingly beneficial effect on increasing the interpretability of deep-net decisions as seen by the visual explanations generated by techniques such as the Grad-CAM. Our code is available at \url{https://github.com/hchoi71/AMC-Loss}.
\end{abstract}
\let\thefootnote\relax\footnotetext{This work was supported in part by NSF CAREER grant number 1452163 and ARO grant number W911NF-17-1-0293.}

\section{Introduction}\label{introduction}

\begin{figure}[ht!]
\begin{center}
\includegraphics[width=0.48\textwidth]{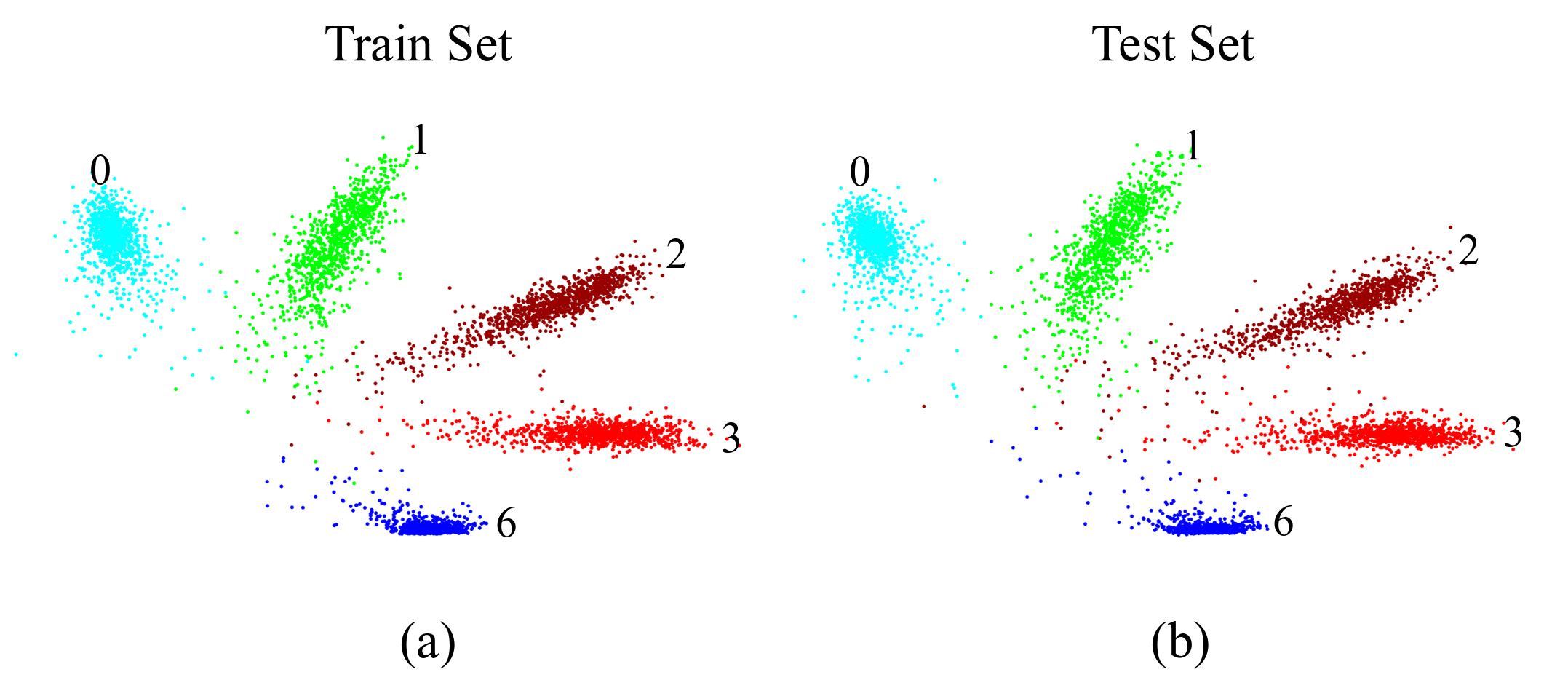}
\end{center}
	\caption{The intrinsic `angular' distribution exhibited in deep features under the cross-entropy loss in MNIST (a) train set, (b) test set where we use $10K/10K$ train/test samples for this visualization. We set the output dimension of the penultimate layer as $2$ and directly plot them in the $2$-dimension space. We select $5$ different classes ($0$, $1$, $2$, $3$, $6$ digits) among $10$ classes and each color denotes a class.}
    \label{fig:angular_distribution}
\end{figure}

Deep learning methods have witnessed great success in solving classification tasks. Especially, the convolutional neural networks (CNNs) have recently drawn a lot of attention in the computer vision community due to its wide range of applications, such as object \cite{he2016deep, he2015delving}, scene \cite{zhou2014object, zhou2014learning}, action recognition \cite{baccouche2011sequential, ji20123d} and so on. The CNN architecture is formed by a stack of convolutional layers which are a collection of filters with learnable weights and biases. Each filter creates feature maps that learn various aspects of an image to differentiate one from the other. In this regard, an essential part of training the networks is the final softmax layer to obtain the predicted probability of belonging to each class. The most common loss function used in the classification task is the cross-entropy loss which computes the cross-entropy over given probability distributions returned by the softmax layer. However, the cross-entropy loss has a few limitations since it only penalizes the classification loss and does not take into account the inter-class separability and intra-class compactness.

\begin{figure}
\begin{center}
\includegraphics[width=.99\linewidth]{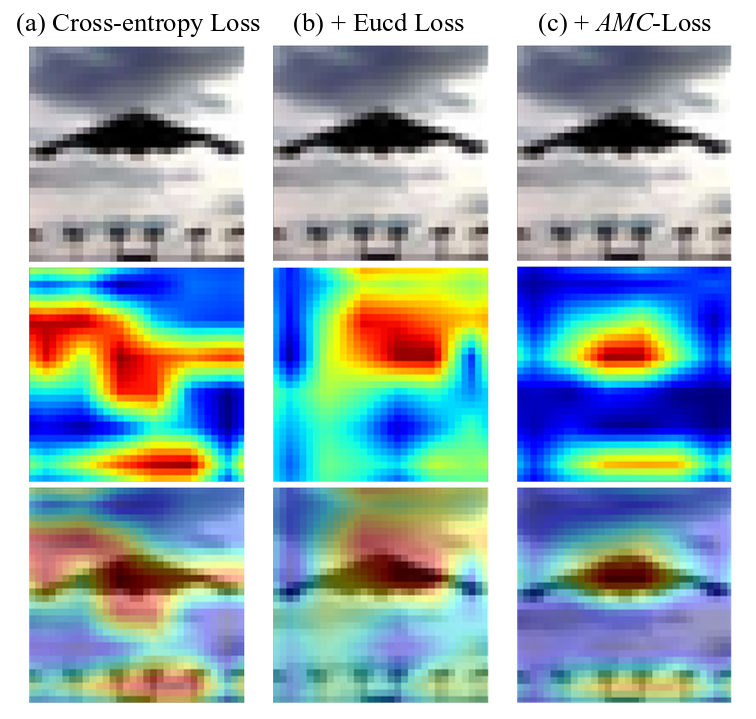}
\end{center}
	\caption{The activation maps generated by the Grad-CAM of the airplane image from different methods, (a) the cross-entropy loss, (b) the Euclidean contrastive Loss, and (c) the {\em AMC}-Loss. The first row indicates the input test image used to generate the activation map in the middle row. The last row shows an overlaid-image with the activation map and the original image. We can observe that {\em AMC}-Loss highlights more discriminative regions while focusing less on the background, leading to more interpretable and explainable models.}
    \label{fig:activation_map}
\end{figure}

To address this issue, many works have gone into utilizing auxiliary loss to enhance the discriminative power of deep features along with cross-entropy loss such as center loss \cite{wen2016discriminative} and contrastive loss \cite{sun2014deep_pred}, where features extracted from the penultimate layer are referred to as deep features in this work. These approaches have greatly improved the clustering quality of deep features. For instance, the center loss which penalizes the Euclidean distance between deep features and their corresponding class centers is a novel technique to enforce extra intra-class distances to be minimized. Even though center loss can improve the intra-class compactness, it would not make distances between different classes not far enough apart, leading to only little changes in inter-class separability. To this end, as an alternative approach, one might want to use contrastive loss in taking into account inter-class separability at the same time. The contrastive loss can be used to learn embedding features to make similar data points close together while maintaining dissimilar ones apart from each other. However, the contrastive loss has to choose a couple of sample pairs to get the loss, so traditional contrastive loss needs careful pre-selection for data pairs (e.g, the neighboring samples/non-neighboring samples). Due to the huge scale of the training set, constructing image pairs inevitably increases the computational complexity, resulting in slow convergence during training. Furthermore, the aforementioned auxiliary loss functions have relied on the Euclidean metric in terms of approximating the semantic similarity of given samples. Meanwhile, studies verified by \cite{liu2017sphereface, wen2016discriminative} hint at the fact that the deep features learned by the cross-entropy loss have an intrinsically `angular' distribution as also depicted in Figure \ref{fig:angular_distribution}, thus seemingly rendering the Euclidean constraint insufficient to combine with the traditional cross-entropy loss. In summary, the weaknesses of the Euclidean contrastive loss include: a) intrinsic mismatch in geometric properties of the learnt features compared to the loss function itself, b) increases the computational complexity by constructing data pairs.

Motivated by these weaknesses, we present the following contributions: \textbf{(1)} First, we introduce a simple representation of images that are mapped as points on a unit-Hilbert hypersphere, which more closely matches the geometric properties of the learnt penultimate features. Also, this results in closed-form expressions to compute the geodesic distances between two points. We are able to directly deploy this geometric constraint into existing contrastive loss formulations. Indeed, our approach can deal with intrinsic angles between deep features on a hyperspherical manifold. \textbf{(2)} Second, designing efficient data pairs is required to compute the geodesic distance between two instances. As a result, we adopt the doubly stochastic sampled data pairs as suggested in \cite{luo2018smooth}, leading to reduced overall computational cost significantly. During training, it does not require much extra time cost with $1$-$2$ seconds per epoch. 

As a preview of the results, we make the surprising finding that the proposed {\em AMC}-Loss results in more explainable interpretations of deep-classification networks, when input activation maps are visualized by the technique of Grad-CAM \cite{selvaraju2017grad}. Generally, the blue regions indicate small or inhibitory weights while the red regions represent large or excitatory weights. Interpreting what parts of the input are most important for the final decision is crucial to make deep-nets more explainable and interpretable. For instance, as seen in Figure \ref{fig:activation_map}, we generated the activation maps from the model trained by three different loss functions, the cross-entropy loss, the Euclidean contrastive loss with cross-entropy loss (refer to this loss as \text{+ Eucd} in the rest of the paper) and {\em AMC}-Loss with cross-entropy loss (denote this by + {\em AMC}-Loss) respectively. In the airplane image, the Euclidean variants seem to be reacting to the body parts of the airplane, but also to the sky. The proposed {\em AMC}-Loss results in a more tightly bounded activation map. Based on additional visualization shown later in section \ref{Experiments}, it appears that the cross-entropy loss pays attention to important parts of objects, but also pays attention to a lot of background information. The Euclidean contrastive loss, when combined with the basic cross-entropy, leads to fuzzy and generally un-interpretable activation maps. Whereas, the addition of the {\em AMC}-Loss results in more compact maps that are also interpretable as distinct object parts, while also reducing the effect of the background.

The rest of the paper is outlined as follows: Section \ref{Background} provides a background study. In Section \ref{Proposed_Method}, we describe the proposed framework in detail and in Section \ref{Experiments} we provide both qualitative and quantitative results. Section \ref{Conclusion} concludes the paper.

\section{Background}\label{Background}
The cross-entropy loss together with softmax activation is one of the most widely used loss functions in image classification \cite{hinton2012improving, he2016deep, he2015delving}. Following this, many joint supervision loss functions with cross-entropy loss have been proposed to generate more discriminative features \cite{wen2016discriminative, sun2014deep_pred, luo2018smooth}. In this section, we focus on revisiting these typical loss functions including related spherical-type loss. 

\paragraph{Cross-entropy Loss} The cross-entropy loss function is defined as:
$
L_{C} = -\frac{1}{N}\sum_{i=1}^{N}\log\frac{e^{W^{T}_{y_{i}}x_{i}+b_{y_{i}}}}{\sum_{j=1}^{n}e^{W^{T}_{j}x_{i}+b_{j}}},
$
where, $x_{i}\in \mathbb{R}^{d}$ denotes the deep features of the $i$-th image, belonging to the $y_{i}$-th class. $W_{j}\in \mathbb{R}^{d}$ represents the $j$-th column of the weights $W\in\mathbb{R}^{d\times n}$ and $b_{j}\in \mathbb{R}^{n}$ is the bias term. The batch size and the class number are $N$ and $n$, respectively. Although the cross-entropy loss is widely used, it does not explicitly optimize the embedding feature to maximize inter-class distance, which results in a performance gap under large intra-class variations. 

\paragraph{Contrastive loss} To address this issue, many related works have attempted to train the network with an auxiliary loss and a cross-entropy loss simultaneously during training. For instance, Yi et al. \cite{sun2014deep} proposed the contrastive loss for face recognition to enforce inter-class separability while preserving intra-class compactness. In specific, one needs to carefully select data pairs to be grouped into neighboring and non-neighboring samples beforehand. That is, if samples belong to the neighbors, the matrix $S_{ij}$ which measures the similarity between samples sets to $1$, otherwise $S_{ij}$ sets to $0$,  meaning samples of different classes. We then denote the contrastive loss as:

\begin{equation}\label{eq:contrastive_loss}
    L_{E} = 
\begin{cases}
    \left\|x_{i}-x_{j}\right\|^{2} & \text{if } S_{ij}=1 \\
    \max(0, m_{e}-\left\|x_{i}-x_{j}\right\|)^{2}, & \text{if } S_{ij}=0
\end{cases}
\end{equation}
where $m_{e}>0$ is a pre-defined Euclidean margin and $ \left\| \cdot \right\|$ is the Euclidean distance between deep features $x_{i}$ and $x_{j}$. Consequently, the neighboring samples are encouraged to minimize their distances while the non-neighboring samples are pushed apart from each other with a minimum distance of $m_{e}$. The value $m_{e}$ is the margin of separation between neighbors and non-neighbors and can be decided empirically. When $m_{e}$ is large, it pushes dissimilar and similar samples further apart thus acting as a margin.

\paragraph{Spherical-type Loss} The existing contrastive loss adopts the Euclidean metric on deep features. However, as mentioned in the previous section, the Euclidean-based loss functions are incompatible with the cross-entropy loss due to intrinsic angular distributions visible in deeply learned features as presented in Figure \ref{fig:angular_distribution}. Weiyang et al. \cite{liu2017sphereface} proposed \textit{SphereFace} to address this issue by introducing angles between deep features and their corresponding weights in a multiplicative way for a face recognition task. For example, for binary class case, the decision boundary for class $1$ and class $2$ become $\left\|\mathbf{x}\right\|(\cos(m\theta_{1})-cos(\theta_{2}))$ and $ \left\|\mathbf{x}\right\|(\cos(\theta_{1})-cos(m\theta_{2}))$ where $m$ quantitatively controls the angular margin and $\theta$ is the angle between weight $\mathbf{W}_{i}$ and feature vector $\mathbf{x}$. Other avenues proposed alternatives to softmax by exploring a spherical family of functions: the spherical softmax and Taylor softmax \cite{de2015exploration}. In spherical softmax, one replaces the exponential function by a quadratic function and the Taylor softmax replaces the exponential function by the second-order Taylor expansion. These alternative formulations allow us to compute exact gradients without computing all the logits, leading to reducing the cost of computing gradients. Although they showed that these functions do not outperform when the length of an output vector is large e.g, in language modeling tasks with large vocabulary size, they surpassed the traditional softmax on MNIST and CIFAR10 dataset. Our work is complementary to these and can be combined with them in that {\em AMC}-Loss intuitively respects the angular distributions empirically observed in deep features.

\begin{figure*}[!]
\begin{center}
\includegraphics[width=0.85\textwidth]{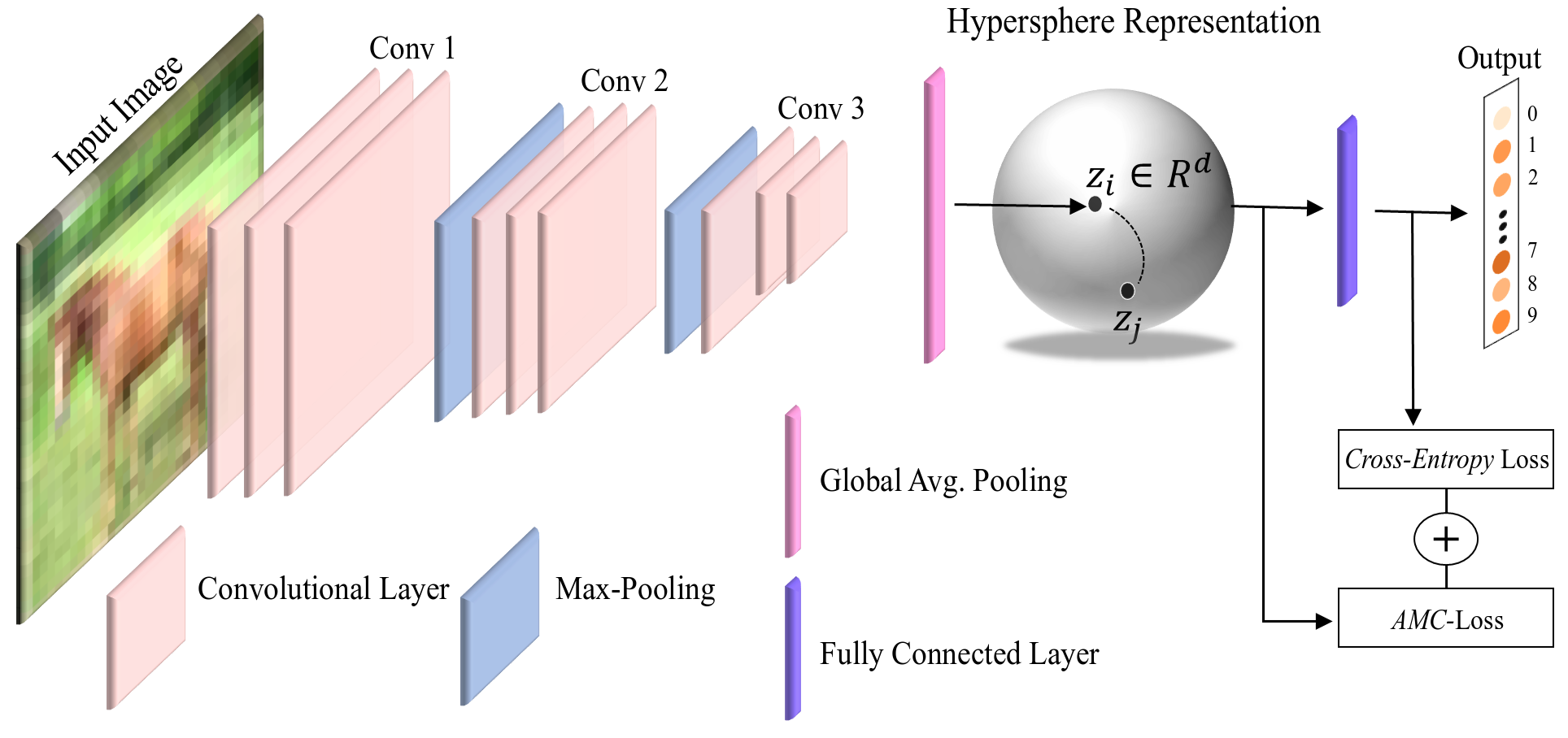}
\end{center}
	\caption{The overall framework of the proposed method. The output of the final convolutional layer is connected to the global average pooling layer whose output dimension is $128$. The fully connected layer with softmax activation outputs the predicted probability of each class which can be used for cross-entropy loss. In the training phase, the deep features $z$ are learned by {\em AMC}-Loss by penalizing geodesic distance between $z_{i}$ and $z_{j}$ and finally, the network is trained with joint-supervision loss e.g, cross-entropy loss and {\em AMC}-Loss. We described the network architecture in detail in Table \ref{network_architecture}.}
    \label{fig:pipeline}
\end{figure*}


\section{Proposed Method}\label{Proposed_Method}
In this section, we elaborate on our approach. A brief overview of the proposed framework for image classification tasks is shown in Figure \ref{fig:pipeline}. The CNNs take the input that is passed through a stack of convolutional (Conv.$1,2,3$) layers, where the filters were used with a small receptive field: $3\times3$. At the last configuration, it utilizes $1\times1$ convolution filters, which can be seen as a linear transformation of the input channels. The max pooling is performed over a $2\times2$ pixel window. The final convolutional layer is then fed to a global average pooling layer, which yields a vector, called a deep feature in this paper. Based on the proposed method, this deep feature is represented as a point on the hypersphere by restricting unit-norm features $z_{i}=x_{i}/\left\|x_{i}\right\|$, and apply them to the {\em AMC}-Loss. The final fully connected layer has a softmax activation function to produce the predicted probability of each class. Finally, the cross-entropy loss function can be used along with {\em AMC}-Loss. That is, during training, our approach estimates a particular embedding position for an image by the hypersphere representations and updates the parameters through the network such that it keeps similar points together and dissimilar ones apart by penalizing the geodesic distance between given two points, $d(z_{i},z_{j})=\cos^{-1}\langle\,z_{i},z_{j}\rangle$. 

\subsection{\textbf{\textit{AMC}}-Loss}\label{ACM_Loss} Intuitively, {\em AMC}-Loss minimizes the geodesic distance for points of the same class while encouraging points of different classes to have a more distinct separation with a minimum angular margin $m_{g}$. To this end, we propose the {\em AMC}-Loss while preserving the existing contrastive loss formulation, as formulated in \eqref{eq:AMC-Loss}.  

\begin{align}\label{eq:AMC-Loss}
    L_{A} = 
\begin{cases}
    (\cos^{-1}\langle\,z_{i},z_{j}\rangle)^{2} & \text{if } S_{ij}=1 \\
    \max(0, m_{g}-\cos^{-1}\langle\,z_{i},z_{j}\rangle)^{2} & \text{if } S_{ij}=0
\end{cases}
\end{align}
where, instead of the Euclidean margin $m_{e}$, $m_{g}>0$ becomes an angular margin. In  \eqref{eq:AMC-Loss}, $S_{ij}=0$ is assigned to non-neighboring pairs whereas $S_{ij}=1$ is allotted to neighboring pairs. Ideally, all possible training sample combinations can be considered in the matrix $S_{ij}$. However, due to the large size of training pairs, instead of updating parameters with respect to the entire training set, we perform the update based on a mini-batch. The CNN model is iteratively optimized using gradient descent by joint-supervision with the cross-entropy loss and the {\em AMC}-Loss. Even though we perform updates at the mini-batch level, it still involves the computational burden to compute the geodesic distance for all combinations of samples in a batch -- thereby resulting in slow-convergence during training. 

Specifically, constructing the matrix $S_{ij}$ involving all data pairs $(x_{i}, x_{j})\in B$ where mini-batch $B$ is of size $n$, requires $\mathcal{O}(n^{2})$ computations in total. Also, computing geodesic distance between two points related to $S_{ij}$ is $\mathcal{O}(p)$ where $p$ denotes $p$-dimensional vector, resulting in the overall computational cost of $\mathcal{O}(n^{2}p)$, which is slow for large $n$. To address this issue, following \cite{luo2018smooth}, we adopt the doubly stochastic sampled data pairs in computing the geodesic. In each iteration, we sample a mini-batch $B$ into two groups, $B_{1}$ and $B_{2}$ groups, where each group has $n/2$ samples. Then we are able to directly compare two groups and compute corresponding geodesic distance element-wise. Finally, the overall cost can be reduced down to $\mathcal{O}(\frac{np}{2})$. Additionally, without the need for pre-selection for data pairs beforehand, we can build the matrix $S_{ij}$ based on the predicted labels from the output of networks $\tilde{f}$ as follows:

\begin{align}\label{eq:similarity_matrix}
    S_{ij} = 
\begin{cases}
    1 & \text{if } \tilde{y}_{i}=\tilde{y}_{j} \\
    0 & \text{if } \tilde{y}_{i}\neq\tilde{y}_{j}.
\end{cases}
\end{align}
The predicted label corresponding to $i$-th input image is given by $\tilde{y}_{i} = \text{argmax}_{k}[\tilde{f}_{i}]_{k}$, where $\text{argmax}_{k}[\cdot]_{k}$ directly indicates the class label having the maximum probability among classes. We present the pseudo-code in algorithm \ref{Alg:mini-batch_training}. 

\begin{algorithm}[h]
\caption{{Mini-batch training of {\em AMC}-Loss} \label{Alg:mini-batch_training}}
\textbf{Input: } $x_{i}$= training inputs, $y_{i}$ corresponding input labels \\
\textbf{Require: } $z_{i}$= normalized deep feature of $x_{i}$ \\
\textbf{Require: } $w(t)$= weight function \\
\textbf{Require: } $f_{\theta}(x)$= CNNs with parameter $\theta$ \\
\begin{algorithmic}[1]
\For{$t$ in $[1, \text{num\_epochs}]$}
\For{each minibatch $B$}
\State $\tilde{f}_{i}$ $\leftarrow$ $f(x_{i}\in B)$
\For{$(x_{i}, x_{j})$ in a minibatch pairs from $B$}
\State compute $S_{ij}$ based on Eq. \ref{eq:similarity_matrix}. 
\EndFor
\State loss $\leftarrow$ $L_{C}$ 
\State \qquad\quad + $w(t)[\lambda\frac{1}{|B|}\sum_{i,j\in B}L_{A}(z_{i}, z_{j}, S_{ij})$
\State update $\theta$ using optimizers (Adam)
\EndFor
\EndFor
\State return $\theta$
\end{algorithmic}
\end{algorithm}

Clearly, the CNNs supervised by {\em AMC}-Loss are trainable and can be optimized by standard SGD. A scalar $\lambda$ denotes the balancing parameter between the cross-entropy loss and $L_{A}$. We additionally conduct experiments to illustrate how the balancing parameter $\lambda$ and angular margin $m_{g}$ influence performance in section \ref{experiment_lambda_margin}. As the weight function $w(t)$, we use the Gaussian ramp-up and ramp-down curve to put weights on $L_{A}$ during training. We describe this weight function in more detail next. 

\subsection{Training Details}\label{implement_training_details}
We implemented our code in Python $3.7$ with Tensorflow $1.12.0$. All models including baseline have been trained for $300$ epochs using Adam optimizer with mini-batch of size $n=128$ and maximum learning rate $0.003$. We use the default Adam momentum parameters $\beta_{1}=0.9$ and $\beta_{2}=0.999$. Following \cite{laine2016temporal}, we ramp up the weight parameter $w(t)$ and learning rate during the first $80$ epoch with weight $w(t)=\exp[-5(1-\frac{t}{80})^{2}]$ and ramp down the learning rate and Adam $\beta_{1}$ to $0.5$ during the last 50 epochs. The ramp-down function is $\exp[-12.5(1-\frac{300-t}{50})^{2}]$. The balancing coefficient of $\lambda$ is set to $0.1$ in all experiments. To compare with the Euclidean contrastive loss, we keep the same architecture and other hyper-parameters settings with $\lambda=0.1$ and $m_{e}=1.0$. The Euclidean margin $m_{e}=1.0$  and angular margin $m_{g}=0.5$ were chosen from different variations, leading to the best performance.

\section{Experiments}\label{Experiments}
In this section, we show the effectiveness of the {\em AMC}-Loss by visualizing deep features in Section \ref{ex:visualization_embedding} and presenting the classification accuracy on several public datasets in Section \ref{classification_results} with a supportive visualization result. Then we investigate the sensitiveness of the balancing parameter $\lambda$ and the angular margin $m_{g}$ in Section \ref{experiment_lambda_margin}. 

\subsection{Improved Clustering}\label{ex:visualization_embedding}

\begin{figure*}
\begin{center}
\includegraphics[width=0.80\linewidth]{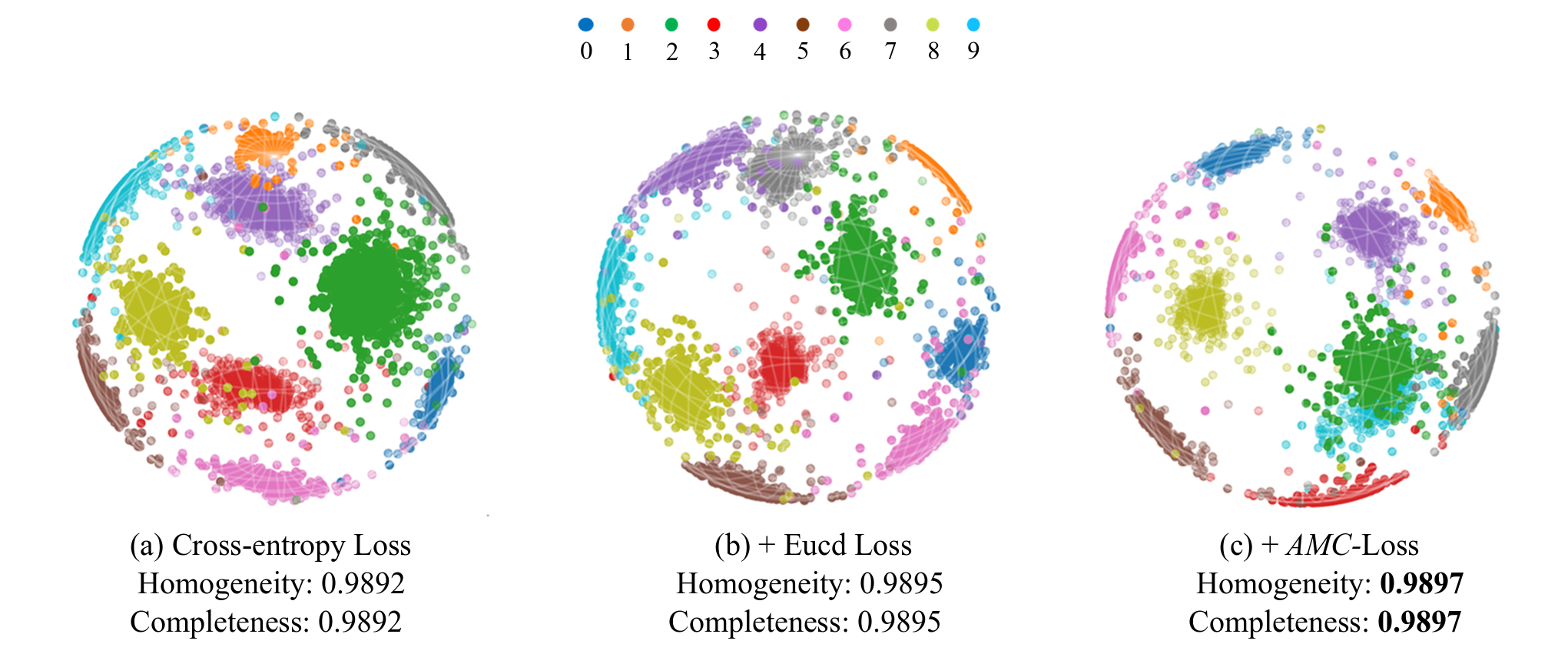}
\end{center}
	\caption{Visualization of features learned with different loss functions on $10K$ MNIST test dataset. We set the output dimension of the penultimate layer as $3$-dimension and then the test samples are directly mapped as points on the unit sphere. Each color denotes a different class. One can see that (a) the cross-entropy loss results in less compactness and separability of learned features whereas (b) the Euclidean contrastive loss (\text{Eucd}) enhances the quality of clustering. (c) Our {\em AMC}-Loss can further increase the intra-class compactness and inter-class separability.}
    \label{fig:sphere_represent}
\end{figure*}

As our proposed framework encourages the deep features to be discriminative on the hyperspherical manifold, we trained the model with the proposed loss function by restricting the feature dimension to three for more intuitive visualization on a sphere. The learnt features are shown in Figure \ref{fig:sphere_represent}. We measure the clustering performance based on the following metrics \cite{rosenberg2007v}:

\textit{\textbf{Homogeneity:}} A clustering result satisfies homogeneity if all of its clusters contain only data points that are members of a single class.

\textit{\textbf{Completeness:}} A clustering result satisfies completeness if all the data points that are members of a given class are elements of the same cluster.

\begin{figure}[h!]
\begin{center}
\includegraphics[width=0.99\linewidth]{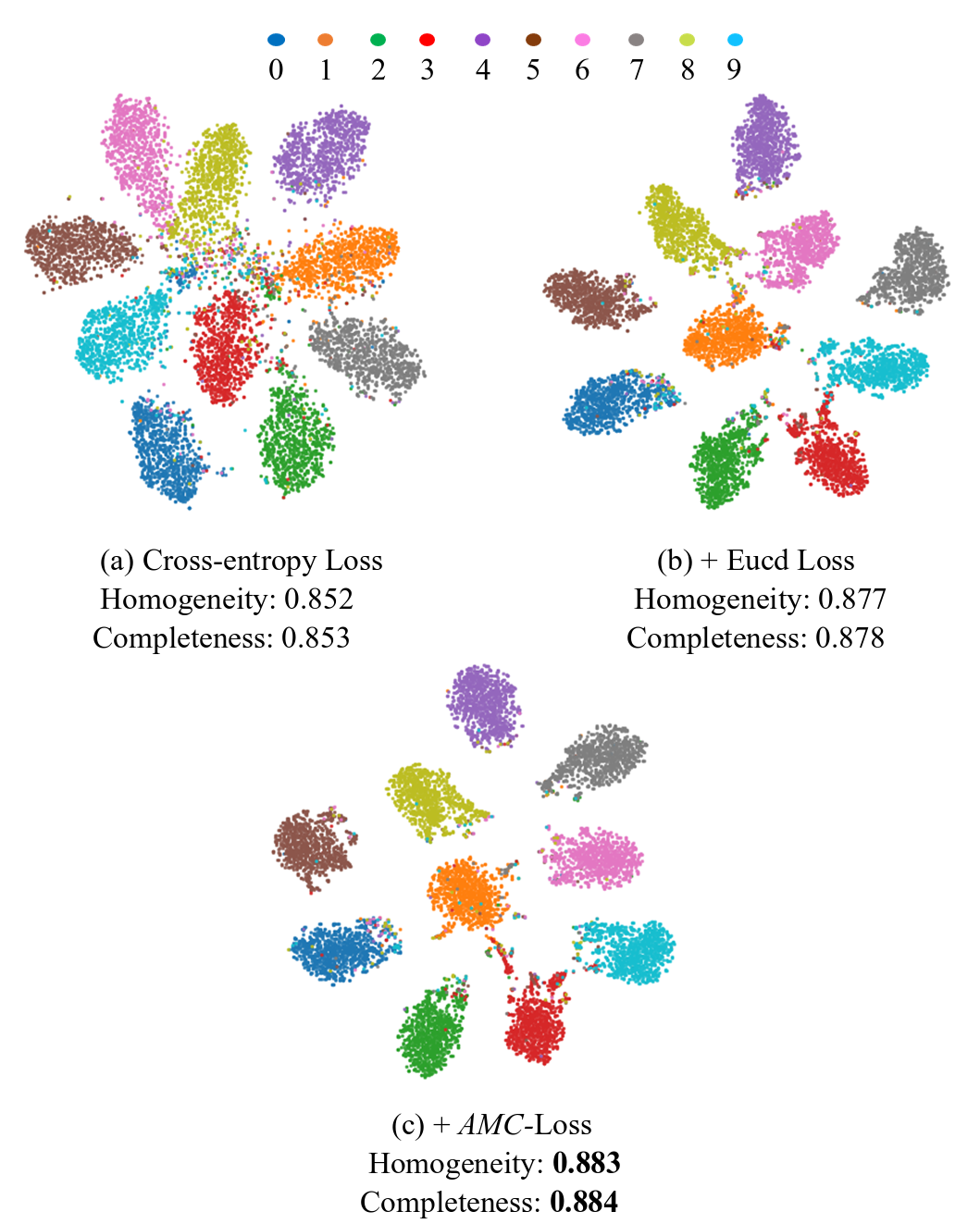}
\end{center}
	\caption{tSNE plot under each model trained with different loss functions on SVHN dataset, (a) cross-entropy loss, (b) \text{+ Eucd} loss and (c) + {\em AMC}-Loss. All points represent deep features projected to $2$-dimension and each color denotes a different class. The {\em AMC}-Loss becomes more distinct feature representations.}
    \label{fig:tSNE}
    \vspace{-0.2in}
\end{figure}

Further, we visualize the deep features learned by {\em AMC}-Loss and compare them with learnt features by baseline models on SVHN test data by projecting them to $2$-dimensions using tSNE \cite{maaten2008visualizing} in Figure \ref{fig:tSNE}. As we can see in this plot, the features learned by + {\em AMC}-Loss are more separable for inter-class samples and more compact for intra-class samples as also seen in homogeneity and completeness.

\subsection{Image Classification}\label{classification_results}
Besides feature representations, we tabulate classification performance on the benchmark dataset in Table \ref{benchmark_result}. The reported results are averaged over $5$ runs. In order to check the significance of the proposed method, we calculate the p-value with respect to only \text{+ Eucd} so that we directly compare the Euclidean constraint with the proposed geometric constraint. The p-value is the area of the two-sided $t$-distribution that falls outside $\pm t$. Although + {\em AMC}-Loss does not yield as good results as the \text{+ Eucd} on MNIST dataset, it outperforms \text{+ Eucd} on other datasets with p-value of less than $0.05$.

\begin{table}[h!]
	\centering
	\scalebox{0.85}{\begin{tabular}{ |c|c|c|c|c| } 
			\hline
			\multirow{2}{*}{ Model} & \multicolumn{2}{c|}{MNIST} & \multicolumn{2}{c|}{CIFAR10} \\ 
			\cline{2-5}
			& Mean$\pm$SD &  p-Value & Mean$\pm$SD & p-Value \\
			\hline
			Cross-entropy & 99.63$\pm$0.01 & - & 82.35$\pm$0.17 & - \\
			+ Eucd & 99.65$\pm$0.01 & - & 82.60$\pm$0.21 & - \\
			+ {\em AMC}-Loss & \textbf{99.66$\pm$0.01} & 0.1525 & \textbf{82.97$\pm$0.20} & 0.0214 \\
			\hline
			\multicolumn{2}{c}{}\\[-0.5em]
			\hline
			\multirow{2}{*}{Model} & \multicolumn{2}{c|}{SVHN} & \multicolumn{2}{c|}{CIFAR100} \\ 
			\cline{2-5}
			& Mean$\pm$SD &  p-Value & Mean$\pm$SD & p-Value \\
			\hline
			Cross-entropy & 94.03$\pm$0.11 & - & 65.16$\pm$0.12 & - \\
			+ Eucd & 95.29$\pm$0.06 & - & 65.57$\pm$0.20 & - \\
			+ {\em AMC}-Loss & \textbf{95.52$\pm$0.05} & 0.0002 & \textbf{66.19$\pm$0.22} & 0.0016 \\
			\hline
			
	\end{tabular}}
	\vspace{0.05in}
	\caption{Classification results on benchmark datasets, averaged over 5 runs. p-values are calculated with respect to the \text{+ Eucd} baseline model.}\label{benchmark_result}
	\vspace{-0.175in}.
\end{table}

\textbf{MNIST.} It consists of the $60,000$ gray-scale training images and $10,000$ test images from handwritten digits $0$ to $9$.

\textbf{CIFAR10.} The CIFAR10 dataset consists of $32\times32$ natural RGB images from $10$ classes such as airplanes, cats, cars and horses and so on. We have $50,000$ training examples and $10,000$ test examples.

\textbf{CIFAR100.} The CIFAR100 dataset consists of $32\times32$ natural RGB images, but they have $20$ super-class (coarse label) and $100$ classes (fine label). Note, we evaluate the performance with $20$ super-class.

\textbf{SVHN.} Each example in SVHN is $32\times32$ color house-number images and we use the official $73,257$ training images and $26,032$ test images.

\begin{figure*}
\begin{center}
\includegraphics[width=0.99\linewidth]{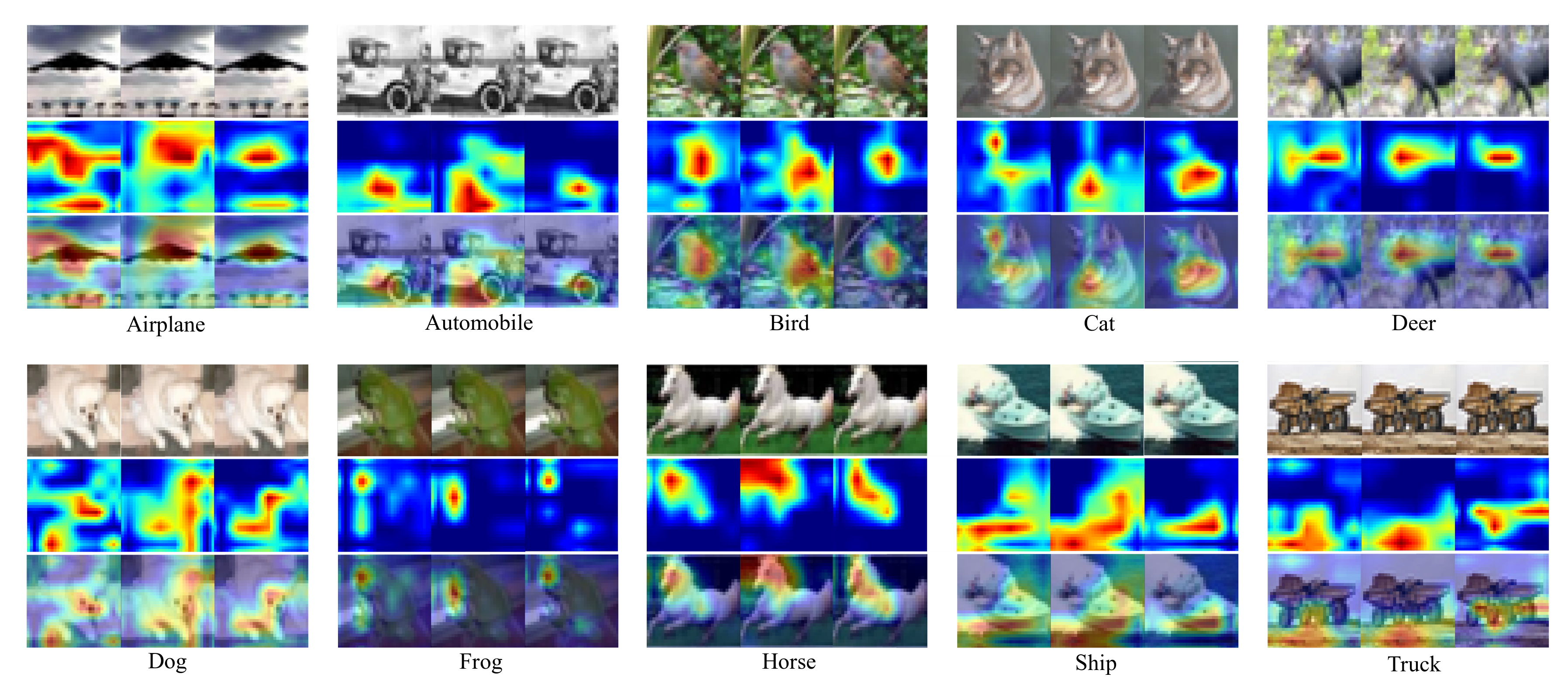}
\end{center}
	\caption{The $3\times 3$ image block set of activation maps generated by the Grad-CAM of test images per class on CIFAR10. The first column of each matrix represents the cross-entropy loss, the \text{+ Eucd} in the second column, and the {\em AMC}-Loss in the third one. We can clearly see the {\em AMC}-Loss highlights the target regions while reducing the background in given images. Particularly, the {\em AMC}-Loss pays more attention to the wheels in the truck image, whereas the cross-entropy and the \text{+ Eucd} seem to mostly react the ground part which shows similar color to the object. Similar observations are seen in the other examples as well.}
    \label{fig:Activation_Map_all}
\end{figure*}

We provide the activation maps of test images per class in Figure \ref{fig:Activation_Map_all}. In this figure, the {\em AMC}-Loss shows qualitatively better performance, in the sense that foreground objects become more distinct and salient with high weights while focusing less on the background. On the other hand, the \text{+ Eucd} loss, including the cross-entropy loss, appears more spread out and less `on target'. By emphasizing important regions, the {\em AMC}-Loss may bring out stronger explainable performance.

\subsection{Robustness to Parameter Tuning}\label{experiment_lambda_margin} 
\begin{figure}[!ht]
\centering
\includegraphics[width=0.95\linewidth]{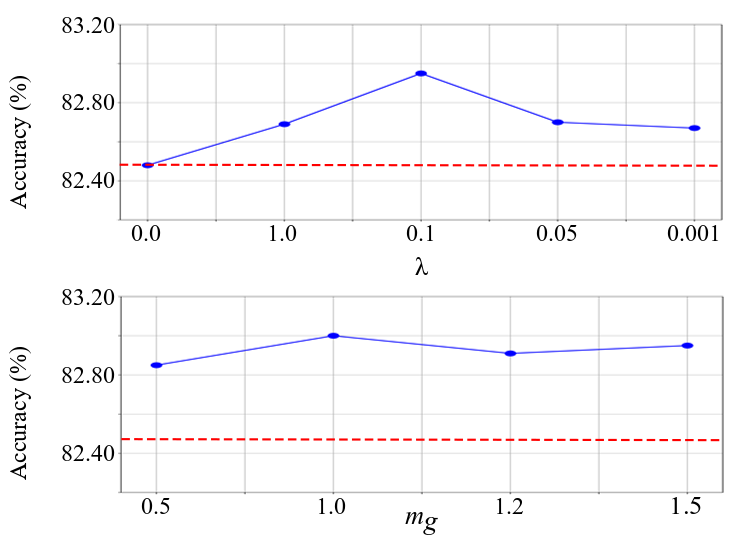}
\begin{tabular}{cc}
\begin{minipage}{.35\linewidth}
\begin{tabular}[b]{cc}\hline
  $\lambda$ & p-Value \\ \hline
  1.0 & 0.1829 \\
  0.1 & 0.0095 \\
  0.05 & 0.2348 \\
  0.001 & 0.2355 \\ \hline
\end{tabular}
\end{minipage} &
\begin{minipage}{.35\linewidth}
\begin{tabular}[b]{cc}\hline
  $m_{g}$ & p-Value \\ \hline
  0.5 & 0.0742 \\
  1.0 & 0.0385 \\
  1.2 & 0.0751 \\
  1.5 & 0.0526 \\ \hline
\end{tabular}
\end{minipage}
\end{tabular}
\vspace{0.05in}
\caption{The average test accuracy on CIFAR10 dataset over $3$ runs, achieved by {\em AMC}-Loss with different $\lambda$ and fixed $m_{g}=0.5$ on the top panel, and with different $m_{g}$ and fixed $\lambda=0.1$ on the bottom one. The red dashed line indicates the baseline model ($\lambda=0$) learned by the cross-entropy loss.}\label{fig:sensitivity_lambda_margin}
\end{figure}

We evaluate our model to see sensitivity to the balancing parameter $\lambda$ and the angular margin $m_{g}$ on the CIFAR10 dataset. The hyper-parameter $\lambda$ controls the balance between the cross-entropy loss and {\em AMC}-Loss and $m_{g}$ determines the minimum angular distance of how far points of non-neighbors are pushed apart. Both of them are essential to our model. We first fix $m_{g}$ to $0.5$ and vary $\lambda$ from $1$ to $0.001$ to learn different models. Likewise, we evaluate the performance by varying angular margin from $0.5$ to $1.5$ with fixed $\lambda=0.1$. The trend accuracy of these models is shown in Figure \ref{fig:sensitivity_lambda_margin} along with the p-value. 

\section{Conclusion}\label{Conclusion}
In this work, we have studied a simple analytic geometric constraint imposed on the penultimate feature layer, motivated by empirical observations of the shape of feature distributions. The proposed {\em AMC}-loss is a more natural way to introduce the contrastive term in combination with the traditional cross-entropy loss function. By representing image features as points on hypersphere manifold, we have shown that deep features learned by the angular metric can enhance discriminative power modestly. More importantly, we find that the use of the {\em AMC}-loss results in models that are seemingly more explainable. This is a rather unexpected finding. As seen in our experiments, the proposed method can enable deep features to be more distinct by improving localization of important regions. Our work is complementary to similar efforts that impose spherical-type losses; the {\em AMC}-Loss extends these ideas into contrastive losses, while showing that the resultant deep-net is more explainable.

\appendix
\section{Network architectures}
\begin{table}[h!]
\centering
\scalebox{0.9}{\begin{tabular}{l c }
 \hline
 Input: $32\times32\times3$ image for CIFAR10, SVHN, CIFAR100 \\ [0.5ex] 
 \hline
 Add Gaussian noise $\sigma$=$0.15$ \\
 $3\times3$ conv. 128 lReLU ($\alpha=0.1$) same padding \\ 
 $3\times3$ conv. 128 lReLU ($\alpha=0.1$) same padding  \\
 $3\times3$ conv. 128 lReLU ($\alpha=0.1$) same padding  \\
 $2\times2$ max-pool, dropout $0.5$  \\
 $3\times3$ conv. 256 lReLU ($\alpha=0.1$) same padding  \\
 $3\times3$ conv. 256 lReLU ($\alpha=0.1$) same padding  \\
 $3\times3$ conv. 256 lReLU ($\alpha=0.1$) same padding  \\
 $2\times2$ max-pool, dropout $0.5$  \\
 $3\times3$ conv. 512 lReLU ($\alpha=0.1$) valid padding  \\
 $1\times1$ conv. 256 lReLU ($\alpha=0.1$) \\
 $1\times1$ conv. 128 lReLU ($\alpha=0.1$) \\
 Global average pool $6\times 6$ $\rightarrow$ $1\times 1$\\ 
 Fully connected $128$ $\rightarrow$ $10$ softmax \\ [1ex]
 \hline
 \hline
 Input: $28\times28\times1$ image for Figure \ref{fig:angular_distribution} and Figure \ref{fig:sphere_represent} \\ [0.5ex] 
 \hline
 Add Gaussian noise $\sigma$=$0.15$ \\
 $3\times3$ conv. 64 lReLU ($\alpha=0.1$) same padding \\ 
 $2\times2$ max-pool, dropout $0.5$  \\
 $3\times3$ conv. 64 lReLU ($\alpha=0.1$) same padding \\ 
 $2\times2$ max-pool, dropout $0.5$  \\
 $3\times3$ conv. 128 lReLU ($\alpha=0.1$) valid padding \\ 
 $1\times1$ conv. 64 lReLU ($\alpha=0.1$) same padding \\ 
 Global average pool $5\times 5$ $\rightarrow$ $1\times 1$\\ 
 Fully connected 3 $\rightarrow$ $10$ softmax \\ [1ex]
 \hline
\end{tabular}}
\vspace{0.05in}
\caption{The network architectures used in all experiments. The output of the global average pooling layer is $128$ for all experiments(exceptions are Figure \ref{fig:angular_distribution} with $2$-dimension and Figure \ref{fig:sphere_represent} with $3$-dimension).}
\label{network_architecture}
\end{table}

As suggested in \cite{laine2016temporal, luo2018smooth}, we use the same CNN architectures for our experiments, but we use standard batch normalization. The architecture in the top panel of Table \ref{network_architecture} is used to produce results for Figures \ref{fig:tSNE},  \ref{fig:activation_map}, and \ref{fig:sensitivity_lambda_margin}, and Table \ref{benchmark_result}. The bottom architecture is used for Figure \ref{fig:angular_distribution} and \ref{fig:sphere_represent}. For the angular distribution in Figure \ref{fig:angular_distribution} and the spherical representation in Figure \ref{fig:sphere_represent}, we trained the model for $150$ epochs using Adam Optimizer with batch size of $128$ and maximum learning rate $0.003$. We apply ramp up during the first $40$ epochs and ramp down for the last $30$ epochs.

{\small
\bibliographystyle{ieee_fullname}
\bibliography{egbib}
}

\end{document}